%% file: main.tex
\documentclass[letterpaper]{article} 
\usepackage{aaai23}  
\usepackage{times}  
\usepackage{helvet}  
\usepackage{courier}  
\usepackage[hyphens]{url}  
\usepackage{graphicx} 
\def\UrlFont{\rm}  
\usepackage{natbib}  
\usepackage{caption} 
\usepackage{algorithm}

\usepackage{newfloat}
\usepackage{listings}
\usepackage{algpseudocode}
\usepackage{amsmath}
\usepackage{booktabs}
\usepackage{color} 
\usepackage{amsfonts}
\usepackage{amssymb}
\usepackage{mathtools}
\usepackage{multirow}
\usepackage{bbm}
\usepackage{dsfont}
\usepackage{colortbl}
\usepackage{subfigure}

\DeclareCaptionStyle{ruled}{labelfont=normalfont,labelsep=colon,strut=off} 
\floatstyle{ruled}
\newfloat{listing}{tb}{lst}{}
\floatname{listing}{Listing}
\title{Relation-Aware Language-Graph Transformer for Question Answering}

\author{
    Jinyoung Park\textsuperscript{\rm 1}\equalcontrib, Hyeong Kyu Choi\textsuperscript{\rm 1}\equalcontrib, Juyeon Ko\textsuperscript{\rm 1}\equalcontrib, Hyeonjin Park\textsuperscript{\rm 2},\\
    Ji-Hoon Kim\textsuperscript{\rm 2,3,4}, Jisu Jeong\textsuperscript{\rm 2,3,4}, Kyungmin Kim\textsuperscript{\rm 2,3,4}, Hyunwoo J. Kim\textsuperscript{\rm 1}\thanks{is the corresponding author} 
}
\affiliations{
    \textsuperscript{\rm 1}Korea University\\ \textsuperscript{\rm 2} NAVER\\ \textsuperscript{\rm 3} NAVER Cloud\\ \textsuperscript{\rm 4} NAVER AI Lab\\

    \{lpmn678, imhgchoi, juyon98, hyunwoojkim\}@korea.ac.kr \\
    \{hyeonjin.park.ml, genesis.kim, jisu.jeong, kyungmin.kim.ml\}@navercorp.com 
}

\usepackage{bibentry}
\input{def}
\input{notation}

\begin{document}

\maketitle

\input{0_Abstract/abstract}
\section{Introduction}
\input{1_Introduction/1_0_introduction}
\section{Related Works}
\input{2_RelatedWorks/2_0_relatedworks}
\section{Preliminaries}
\input{3_Method/preliminaries}
\section{Question Answering Transformer}
\input{3_Method/3_0_method}
\section{Experiments}

\input{4_Experiments/4_0_experiments}
\section{Analysis} 
\label{sec:analysis}
\input{5_Analysis/5_0_analysis}
\section{Conclusion}
\input{6_Conclusion/conclusion}

\section*{Acknowledgments}
This work was partly supported by NAVER Corp. and MSIT (Ministry of Science and ICT), Korea, under the ICT Creative Consilience program~(IITP-2023-2020-0-01819) supervised by the IITP.



{
\small
\bibliography{aaai23}
}

\input{appendix}
\end{document}

%% file: def.tex
\newcommand{\omitme}[1]{}

\newcommand{\Oc}{\mathcal{O}}

\def\eg{\emph{e.g}.}
\def\ie{\emph{i.e}.}

\def\Gc{\mathcal{G}}

\def\Nc{\mathcal{N}}

\def\Nc{\mathcal{N}}

\makeatletter
\newcount\my@repeat@count
\newcommand{\myrepeat}[2]{%
  \begingroup
  \my@repeat@count=\z@
  \@whilenum\my@repeat@count<#1\do{#2\advance\my@repeat@count\@ne}%
  \endgroup
}
\makeatother

%% file: notation.tex
\newcommand{\graph}{\Gc}

\newcommand{\nodeset}{\mathcal{V}}
\newcommand{\edgeset}{\mathcal{E}}
\newcommand{\node}{v}

\newcommand{\relset}{\mathcal{R}}
\newcommand{\nodetypeset}{\mathcal{T}_v}
\newcommand{\rel}{r}
\newcommand{\question}{q}
\newcommand{\answer}{a}

\newcommand{\aset}{\mathcal{C}}
\newcommand{\lmin}{\boldsymbol{x}}
\newcommand{\lminset}{\mathcal{X}}
\newcommand{\cls}{[\text{cls}]}
\newcommand{\sep}{[\text{sep}]}
\newcommand{\lmout}{\boldsymbol{h}^{\text{LM}}}
\newcommand{\kgout}{\boldsymbol{h}}

\newcommand{\lmoutset}{\mathcal{H}}
\newcommand{\kgoutset}{\mathcal{H}}
\newcommand{\jointoutset}{\mathcal{Z}}

\newcommand{\pathset}{\boldsymbol{\Psi}}
\newcommand{\paths}{\psi}

\newcommand{\nodefeat}{\boldsymbol{f}}
\newcommand{\edgeencoder}{g_{\theta_1}}
\newcommand{\pathencoder}{g_{\theta_2}}
\newcommand{\head}{h}
\newcommand{\tail}{t}
\newcommand{\interm}{v}

\newcommand{\rpe}{\Omega}
\newcommand{\rpescalar}{\omega}
\newcommand{\neighbor}{\Nc}
\newcommand{\mes}{\boldsymbol{m}}
\newcommand{\typeembed}{\boldsymbol{e}}
\newcommand{\srctoken}{\boldsymbol{X}}
\newcommand{\querytoken}{\boldsymbol{Q}}
\newcommand{\keytoken}{\boldsymbol{K}}
\newcommand{\valuetoken}{\boldsymbol{V}}
\newcommand{\loss}{\mathcal{L}}
\newcommand{\numheads}{H}
\newcommand{\numlayers}{L}
\newcommand{\onehot}{\phi}

%% file: 0_Abstract/abstract.tex
Question Answering~(QA) is a task that entails reasoning over natural language contexts, and many relevant works augment language models~(LMs) with graph neural networks~(GNNs) to encode the Knowledge Graph~(KG) information.
However, most existing GNN-based modules for QA do not take advantage of rich relational information of KGs and depend on limited information interaction between the LM and the KG.
To address these issues, we propose Question Answering Transformer~(QAT), which is designed to jointly reason over language and graphs with respect to entity relations in a unified manner.
Specifically, QAT constructs Meta-Path tokens, which learn relation-centric embeddings based on diverse structural and semantic relations.
Then, our Relation-Aware Self-Attention module comprehensively integrates different modalities via the Cross-Modal Relative Position Bias, which guides information exchange between relevant entites of different modalities.
We validate the effectiveness of QAT on commonsense question answering datasets like CommonsenseQA and OpenBookQA, and on a medical question answering dataset, MedQA-USMLE.
On all the datasets, our method achieves state-of-the-art performance. 
Our code is available at http://github.com/mlvlab/QAT.

%% file: 1_Introduction/1_0_introduction.tex
Question Answering~(QA) is a challenging task that entails complex reasoning over natural language contexts.
The large-scale pre-trained language models~(LMs) have proven to be capable of processing such contexts, by encoding the implicit knowledge in textual data.
To deal with various QA tasks, both generic and domain-specific LMs~\cite{liu2019roberta,lee2020biobert,liu2020self} have been employed.
But the QA systems only with LMs often struggle with structured reasoning due to the lack of explicit and structured knowledge. 
Thus, many works~\cite{mihaylov2018knowledgeable,lin2019kagnet,feng2020scalable,wang2021gnn,yasunaga2021qa,zhang2022greaselm} in the literature have studied to effectively incorporate Knowledge Graphs~(KGs) into the question-answering system.

Most existing efforts for fusing LM and KG utilized a separate graph neural network~(GNN) module to process the KG and integrated it with LM in the `final' prediction step~\cite{feng2020scalable,yasunaga2021qa,wang2021gnn}.
However, recent works pinpointed the shortcomings of the shallow fusion, and have proposed methods that fuse LM and KG early in the intermediate layers.
They aimed to find more appropriate ways to exchange information between the two modalities, via special tokens and nodes~\cite{zhang2022greaselm} or cross-attention~\cite{sun2021jointlk}.
But the approaches have a modality-specific encoder GNN and the information exchange occurs only on fusion layers resulting in modest information exchange.
In addition, the GNNs in previous works are {\em not} suited for fully exploiting KG data, which comprise triplets of two entities and their relations.
To process the rich relational information, a more \textit{relation-centric} KG processor is necessary, in place of the \textit{node-centric} GNNs that learn node features with the message-passing scheme.

To this end, we propose the \textbf{Question Answering Transformer~(QAT)}, specifically designed to jointly reason over the LM and KG in a unified manner, without an explicit KG encoder, \eg, GNN.
We encode relation-centric information of the KG in the form of Meta-Path tokens.
Along with LM outputs, the Meta-Path tokens are input into our Relation-Aware Self-Attention~(RASA) module, whose cross-modal information integration is facilitated by our Cross-Modal Relative Position Bias.
The diverse structural and semantic relations along the meta-paths between question-answer entities can be easily integrated with the LM context via Meta-Path Tokens.
Also, the Meta-Path token sets tend to be diverse, bringing enhanced expressivity in self-attention.
Moreover, we present a method to exchange information in a unified manner, guided by the Cross-Modal Relative Position Bias.
This provides a more flexible means to fuse LM and KG, encouraging information exchange between relevant entities across different modalities.

Then, our \textbf{contributions} are as follows:
\begin{itemize}
    \item We introduce the Meta-Path token, a novel embedding to encode KG information based on diverse relations along meta-paths.
    \item We present Cross-Modal Relative Position Bias, which enables more flexible information exchange encouraging interactions between relevant entities across different modalities: LM and KG.
    \item We propose Question Answering Transformer, a relation-aware language-graph transformer for question answering by joint reasoning over LM and KG.
    \item We achieve state-of-the-art performances on commonsense question answering: CommonsenseQA and OpenBookQA; and on medical question answering: MedQA-USMLE, thereby demonstrating our method's joint reasoning capability.
\end{itemize}

%% file: 2_RelatedWorks/2_0_relatedworks.tex
\paragraph{Knowledge Graph based QA.}
KGs are structured data containing relational information on concepts~(\eg, ConceptNet~\cite{speer2017conceptnet}, Freebase~\cite{bollacker2008freebase}, and Wikidata~\cite{vrandevcic2014wikidata}), and they have been widely explored in the QA domain.
Recent studies on KG based QA process the KG in an end-to-end manner with graph neural networks~\cite{feng2020scalable,yasunaga2021qa,wang2021gnn} or language models~\cite{wang2020connecting,bansal2022cose} for knowledge module. 
Most of them focus on adequately processing the KG first, while independently training the LM in parallel.
The two modalities are then fused in the final step to render a QA prediction.
But this shallow fusion does not facilitate interactions between the two modalities, and several methods~\cite{sun2021jointlk,zhang2022greaselm} to fuse LM and KG in the earlier layers have recently been introduced as well.
The commonality of these works is that the KG nodes are processed with the GNN message passing scheme, which are then fused with LM embeddings.
Contrary to this line of works, our method manages the joint reasoning process by explicitly learning the multi-hop relationships, and facilitating active cross-modal attention.

\paragraph{Multi-relational Graph Encoder.}
GNNs have shown superior performance on various type of graphs in general~\cite{kipf2017semi,velivckovic2017graph}.
However, additional operations may be required to encode the rich relational information of multi-relational graphs like the KGs. 
The KG is a directed graph whose edges are multi-relational, and several relevant works~\cite{schlichtkrull2018modeling,wang2019heterogeneous} have been introduced to process such graph types.  
Unlike homogeneous graph neural networks such as the GCN~\cite{kipf2017semi}, exploiting rich semantic information is essential to encoding multi-relational graphs like KGs.
For instance, R-GCN~\cite{schlichtkrull2018modeling} performs graph convolution for each relation to consider the different interactions between entities. 
HAN~\cite{wang2019heterogeneous} adopted meta-paths to enrich semantic information between nodes by defining path-based neighborhoods. 
Moreover, \cite{hwang2020self} has proposed meta-path prediction as a self-supervised learning task to encourage GNNs to comprehend relational information.

%% file: 3_Method/preliminaries.tex
\paragraph{Question Answering with Knowledge Graphs.}
In this paper, we consider the question answering task via joint reasoning over the LM and KG.
The objective of the task is to understand the context of the given question by jointly reasoning on the language and structured relational data.
Specifically, we use masked LMs (\eg, RoBERTa~\cite{liu2019roberta}, SapBERT~\cite{liu2020self}) and multi-relational KGs (\eg, ConceptNet).

In multiple-choice question answering~(MCQA), word entities in question $\question$ and answer choice $\answer\in\aset$ are concatenated to form an input sequence $\lminset = \left[\cls, \lmin_{\question_1}, \ldots, \lmin_{\question_{|\question|}}, \sep, \lmin_{\answer_1}, \ldots, \lmin_{\answer_{|\answer|}}\right]$.
Then, a pretrained language model $g_\text{LM}$ computes the context tokens $\lmoutset_\text{LM}$ as $\lmoutset_\text{LM} = g_\text{LM}(\lminset)$.
To enable joint reasoning with external knowledge, we leverage the KG containing structured relational data.
Following prior works~\cite{lin2019kagnet, yasunaga2021qa, wang2021gnn}, we extract a multi-relational subgraph $\graph = (\nodeset, \edgeset)$ from the full KG for each question and answer choice pair.
We denote $\nodeset$ as a set of nodes~(entities) and $\edgeset \subseteq \nodeset \times \relset \times \nodeset$ as the set of subgraph edges where $\relset$ indicates the edge types between entities.
Also, we respectively denote $\nodeset_\question \subseteq \nodeset$ and $\nodeset_\answer \subseteq \nodeset$ as the set of nodes mentioned in the given question~$\question$ and answer choice~$\answer$.
Given the multi-relational subgraph, a graph encoder $g_\text{KG}$ encodes $\graph$ as $\kgoutset_\text{KG} = g_\text{KG}(\graph)$. 
Given multimodal information $\lmoutset_\text{LM}$ and $\kgoutset_\text{KG}$, module $\mathcal{F}$ can be applied to output a joint representation $\jointoutset_{\answer\in\aset} = \mathcal{F}(\lmoutset_\text{LM}, \kgoutset_\text{KG})$.
Here, $\mathcal{F}$ can be as simple as feature concatenation or as complex as the Transformer module~\cite{vaswani2017attention}.

\paragraph{Meta-Paths.}
The meta-path~\cite{sun2011pathsim} is defined as a path on a graph $\graph$ whose edges are multi-relational edges, \ie, $v_1 \xrightarrow[]{r_1} v_2 \xrightarrow[]{r_2} \ldots \xrightarrow[]{r_l} v_{l+1}$, where $r_l \in \mathcal{R}$ refers to the edge type of the $l^\text{th}$ edge within the meta-path.
Then, we may view the meta-path as a composite relation $\mathbf{r} = r_1 \circ r_2 \circ \ldots \circ r_l$ that defines the relationship between nodes $v_1$ and $v_{l+1}$.
The meta-path can be naturally extended to the multi-hop relation encoding, providing a useful means to define the relationship between an arbitrary entity pair from the KG subgraph.
For instance, a meta-path (jeans -- \textit{related\_to} $\rightarrow$ merchants -- \textit{at\_location} $\rightarrow$ shopping\_mall) can be defined to describe the relationship between `jeans' and `shopping mall' for the question in Figure~\ref{fig:choice}.

%% file: 3_Method/3_0_method.tex
In this section, we introduce Question Answering Transformer (QAT).
QAT is a relation-aware language-graph Transformer for knowledge-based question answering, which performs joint reasoning over LM and KG.
We take a novel approach of encoding the \textit{meta-path} between KG entities to jointly self-attend to KG and LM representations.
This is enabled by our Meta-Path tokens (Section~\ref{subsection:mptoken}) and our Relation-Aware Self-Attention (Section~\ref{subsection:rasa}) module.

\input{3_Method/methods}

%% file: 3_Method/methods.tex
\input{Figures/tex/main}

\subsection{Meta-Path Token Embeddings}
\label{subsection:mptoken}
The gist of question answering is to understand the relationship between entities.
Accordingly, QAT learns embeddings that represent the relationship between each node pair, dubbed Meta-Path tokens.

When encoding the edge feature, the edge type $\rel \in \relset$ is utilized in the form of a one-hot vector.
The node types are also provided as input, but the node features are not directly employed when learning the embeddings.
Following~\citeauthor{yasunaga2021qa}, a node type among $\nodetypeset = \{\mathbf{Z}, \mathbf{Q}, \mathbf{A}, \mathbf{O}\}$ is selected based on the node's source.
To elaborate, $\mathbf{Z}$ is a node which is deliberately inserted to connect all question and answer entities for context aggregation (similar to the \textit{cls} token in Transformers), $\mathbf{Q}$ and $\mathbf{A}$ each denote the entities from the question and answer choice, respectively. $\mathbf{O}$ is the ``other'' type of node that does not fall in the three categories, but constitutes the extracted subgraph. 
We further input the feature translation~\cite{bordes2013translating} of the head~($\head$) and tail~($\tail$) nodes that comprise an edge  $\head \xrightarrow[]{r} \tail$.
We compute the translation $\delta_{\head,\tail}$ as $\delta_{\head, \tail} = \nodefeat_\tail - \nodefeat_\head,$
where $\nodefeat_\tail$, $\nodefeat_\head$ are the tail and head node features, respectively.
Then, the embedding for an edge is computed as 
\begin{equation}
    \kgout_{(\head, \rel, \tail) \in \edgeset} = \edgeencoder([\onehot(\head), \rel, \onehot(\tail), \delta_{\head,\tail}]),
\end{equation}
where $\edgeencoder(\cdot)$ is a MLP and $\onehot(\cdot):\nodeset \rightarrow \nodetypeset$ is a one-hot encoder that maps a node $\node \in \nodeset$ to a node type in $\nodetypeset$. 

In addition, we aim to diversify the embedding set by incorporating multi-hop paths.
Specifically, we learn the embedding for each path $\paths$ between a question node and an answer node; either one is the head or tail node.
We consider paths of length $n$, and utilize a separate network for each $n$-hop path embeddings, where $n$-hop path $\paths$ from $\head$ to $\tail$ is featurized as $[\onehot(\head), \rel_1, \onehot(\interm_1), \ldots, \onehot(\interm_{n-1}), \rel_{n}, \onehot(\tail), \delta_{\head,\tail}]$.
Then, for instance, the Meta-Path token of a 2-hop path is
\begin{equation}
    \kgout_{\paths \in \pathset_2} = \pathencoder([\onehot(\head), \rel_1, \onehot(\node_1), \rel_2, \onehot(\tail), \delta_{\head,\tail}]),
\end{equation}
where $\pathencoder$ denotes the 2-hop path encoder and $\node_1$ is an intermediate node.
$\pathset_2$ refers to the set of all 2-hop paths that connects $\head$ and $\tail$.
Since an edge can be interpreted as a 1-hop path, we may concisely define the Meta-Path token set as
\begin{equation}
    \kgoutset_\text{KG} = \bigcup_{k=1}^K \{\kgout_\paths | \paths \in \pathset_k\},
    \label{eq:mptokenset}
\end{equation}
where we set $K$ to 2 in our experiments.

Compared to QAT, many previous works have employed graph neural networks to encode the KG~\cite{yasunaga2021qa, wang2021gnn, lin2019kagnet}.
Generally adopting the message-passing mechanism, the neighboring nodes and edge features are aggregated by propagating a layer, and the \textit{nodes} are finally pooled for prediction.
Although the nodes consequently incorporate relational information, the objective of message-passing is to learn node features rather than relations, characterizing GNNs to be innately \textit{node-centric}.
On the other hand, our QAT explicitly encodes the relational information into Meta-Path tokens.
By utilizing Meta-Path tokens for joint reasoning, the model takes advantage of the \textit{relation-centric} tokens which contain information on the structural and semantic relations within the subgraph, see Section~\ref{subsection:discussions} for further discussions.
%



\paragraph{Drop-MP.}
Inspired by \cite{rong2019dropedge}, we selectively apply a simple tactic of stochastically dropping meta-paths for regularization.
Considering that each meta-path corresponds to one Meta-Path token, we simply drop a certain portion of Meta-Path tokens during training.


\subsection{Relation-Aware Self-Attention}
\label{subsection:rasa}
After retrieving the LM token embeddings $\lmoutset_\text{LM}$ and Meta-Path tokens $\kgoutset_\text{KG}$ (Eq.~\eqref{eq:mptokenset}), we now jointly reason over the concatenated token set to compute a logit value for each question and answer choice pair.

\paragraph{Language-Graph Joint Self-Attention.}
The multi-head self-attention module in the Transformer model is well known for its learning capacity and flexibility.
By concatenating the two modalities and applying self-attention, each token aggregates features based on inter- and intra-modal relations.
We also add learnable modality embeddings, $\typeembed_\text{LM}$ and $\typeembed_\text{KG}$, to the query and key tokens in order to encode the modality source into feature vectors.
Then, the self-attention output $\hat{\jointoutset}$ is computed by
\begin{equation}
    \srctoken = [\lmoutset_\text{LM}; \kgoutset_\text{KG}], \hspace{2mm} 
    \tilde{\srctoken} = [\lmoutset_\text{LM} + \typeembed_\text{LM}; \kgoutset_\text{KG} + \typeembed_\text{KG}]
\end{equation}
\begin{equation}
    \querytoken = \tilde{\srctoken} W_{\querytoken}, \hspace{2mm}
    \keytoken = \tilde{\srctoken} W_{\keytoken}, \hspace{2mm}
    \valuetoken = \srctoken W_{\valuetoken}
\end{equation}
\begin{equation}
    \hat{\jointoutset} = \text{Softmax}( \querytoken  \keytoken^\top / \sqrt{d}) \hspace{1mm} \valuetoken,
\end{equation}
where $W_{\querytoken}$, $W_{\keytoken}$, and $W_{\valuetoken}$ refers to the linear projection weights, and $d$ is the key embedding dimension.
Note, we described a single-headed case to avoid clutter.

\input{Figures/tex/rpe}

\paragraph{Cross-Modal Relative Position Bias.}
Although the Transformer architecture alone is powerful in itself, providing guidance on learning \textit{how} to aggregate information further enhances performance~\cite{liu2021swin, yang2021focal, wu2021rethinking}.
To encourage joint reasoning, we utilize a learnable relative position bias $\rpe$ that controls the cross-modal attention weights between relevant tokens.

The interrelation of a token pair from the two modalities is determined by whether the natural language representation of each entity is similar to each other.
We utilize the pretrained GloVE~\cite{pennington2014glove} word embeddings to vectorize each entity.
In the case of KG, several node entities contain a series of words joined with underbars (\eg, jewelry\_store).
For those entities, we take the average of the discretized word's GloVE embedding as its representation vector (\eg, [GloVE(``jewelry'') + GloVE(``store")] / 2).
Then, for each relation token in $\kgoutset_\text{KG}$, we take its pertinent head and tail entity to compute their similarity with LM entities using cosine similarity, respectively.
The head and tail nodes of a Meta-Path token are mapped to an LM token in $\lmoutset_\text{LM}$ by selecting the LM token with the highest similarity.
That is, each Meta-Path token embedding is matched with two LM tokens.
Based on this matching, our Cross-Modal Relative Position Bias~(RPB) $\rpe \in \mathbf{R}^{|\srctoken|\times|\srctoken|}$ is defined.
With abuse of notation, $i$, $j$ refer to the $i^\text{th}$, $j^\text{th}$ token:

Let
\begin{equation*}
\small
\begin{split}
\mathbf{p}_{ij} &= \mathds{1} \left[i= \underset{i' \in \lmoutset_\text{KG}}{\text{argmax}} f(i'_v)\cdot f(j), \exists v \in \{h, t\}\right]\cdot \mathds{1}\left[ j \in \kgoutset_\text{LM}\right],\\
\mathbf{q}_{ij} &= \mathds{1} \left[i = \underset{i' \in \lmoutset_\text{LM}}{\text{argmax}} f(i')\cdot f(j_v), \exists v \in \{h, t\}\right] \cdot \mathds{1}\left[ j \in \lmoutset_\text{KG} \right].\\
\end{split}
\end{equation*}
Then,
\begin{equation}
    \rpe[m, n] = 
    \begin{cases}
        \hspace{2mm} \rpescalar_1  & \text{, if} \hspace{3mm} \mathbf{p}_{mn} = 1, \mathbf{q}_{mn} = 0,\\
        \hspace{2mm} \rpescalar_2  & \text{, if} \hspace{3mm} \mathbf{p}_{mn} = 0, \mathbf{q}_{mn} = 1,\\
        \hspace{2mm} 0  & \text{, otherwise} \\
    \end{cases}
\end{equation}

where $f(\cdot)$ is the GloVE-based embedding, $i_v$ is the text representation of node $v$ from token $i$, and $\mathds{1}[\cdot]$ is the indicator function.
$\rpescalar_1$, $\rpescalar_2$ are learnable scalar parameters, and $m$ and $n$ are token indices of $\srctoken$. 
Also note that $\rpe$ is separately defined for each attention head, see Figure~\ref{fig:rpe}.

Then, our Relation-Aware Self-Attention~(RASA) can be finally written as
\begin{equation}
\label{eq:rasa}
    \hat{\jointoutset} = \text{Softmax}(\querytoken  \keytoken^\top / \sqrt{d} + \rpe) \hspace{1mm} \valuetoken,
\end{equation}
which is followed by an FFN layer that aggregates multiple heads.
To enlist our transformer layer computation :
\begin{equation}
    \begin{split}
        \hat{\jointoutset} &= \text{RASA}(\text{LN}(\srctoken)) + \srctoken \\
        \jointoutset &= \text{FFN}(\text{LN}(\hat{\jointoutset})) + \hat{\jointoutset},
    \end{split}    
\end{equation}
where LN is the layer norm, and RASA refers to our Relation-Aware Self-Attention module (Eq.~\eqref{eq:rasa}).

Furthermore, to encourage a positive bias, we use a regularizer term in the final loss as
\begin{equation}
    \loss = \loss_{\text{CE}} - \lambda \hspace{1mm} \sum_{l=1}^\numlayers\sum_{h=1}^\numheads 
    \sigma(\rpescalar^{(hl)})
\end{equation}
where $\loss_\text{CE}$ is the cross entropy loss for question answering, and $\sigma$ can be a nonlinear function (\eg, logarithm, tanh). 
$\rpescalar^{(hl)}$ refer to the relative position bias terms from the $h^{th}$ attention head of layer $l$, and $\lambda$ is a constant.

\subsection{Comparisons with Existing Methods}
\label{subsection:discussions}
\input{3_Method/discussions.tex}

%% file: Figures/tex/main.tex
\begin{figure*}[ht!]
\begin{center}
\includegraphics[width=1\textwidth]{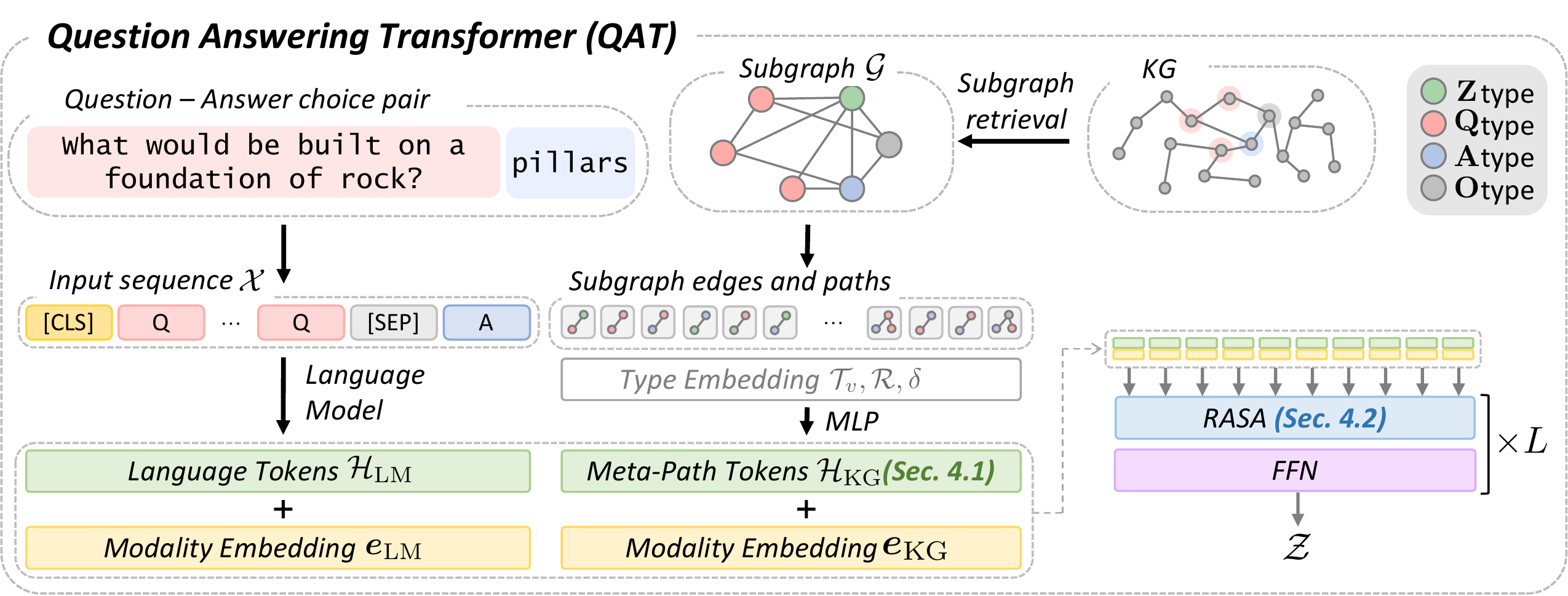}
\end{center}
\caption{Overall architecture of QAT. The Question Answering Transformer~(QAT) is a relation-aware language-graph transformer for knowledge-based question answering tasks. 
Given  a question-answer choice pair, a relevant subgraph $\graph$ is retrieved from the knowledge graph.
Then, the Meta-Path token is encoded for each $n$-hop meta-path with type embeddings based on structural and semantic relationships between nodes (Section~\ref{subsection:mptoken}). 
Then, the set of Meta-Path tokens $\kgoutset_\text{KG}$ are concatenated with the LM outputs $\lmoutset_\text{LM}$, to be fed into our Relation-Aware Self-Attention module. 
The Relation-Aware Self-Attention module effectively aggregates multimodal information via the Cross-Modal Relative Position Bias (Section~\ref{subsection:rasa}). 
After $L$ attention blocks, our QAT renders $\jointoutset$ for each answer choice, which is later passed through a prediction head to output a logit value.}
\label{fig:mainfig}
\end{figure*}

%% file: Figures/tex/rpe.tex
\begin{figure}[t!]
\begin{center}
\includegraphics[width=1.0\columnwidth]{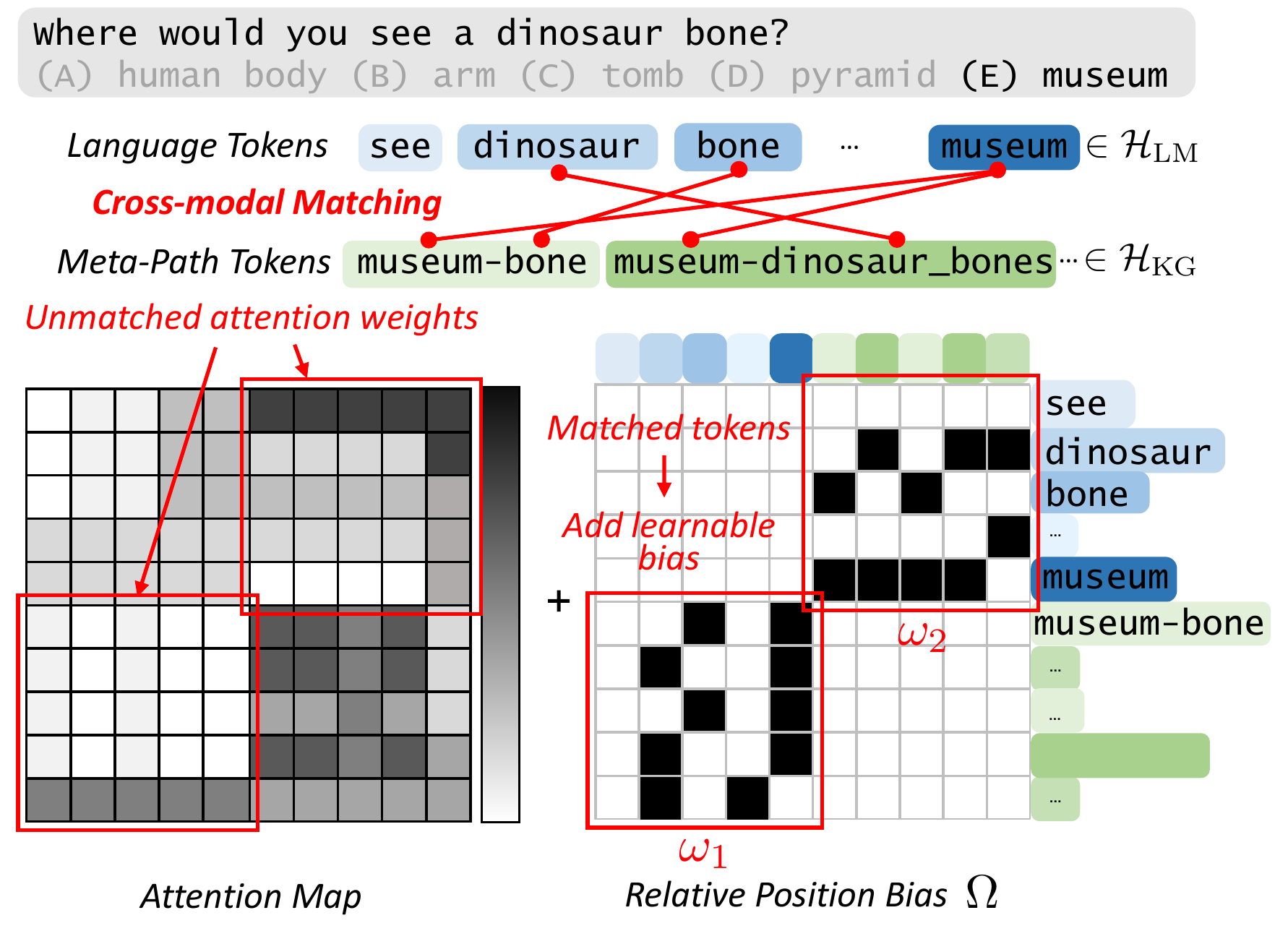}
\end{center}
\caption{Illustration of Cross-Modal Relative Position Bias. Each Meta-Path token incorporates at least one head $\head$ and tail $\tail$ node. The two nodes are each matched with an LM token, whose corresponding position is parameterized in our relative position bias $\rpe$. Then, $\rpe$ is added to the attention map (Eq.~\eqref{eq:rasa}).}
\label{fig:rpe}
\end{figure}

%% file: 3_Method/discussions.tex
In this section, we discuss QAT with GNN-based methods and RN-based methods for question answering problems.  

\paragraph{GNN-based methods.}
GNNs learn node representations with the message-passing scheme, which recursively aggregates the representations of neighbors as follows~\cite{gilmer2017neural,kipf2017semi,yasunaga2021qa}:
\begin{equation}
    \kgout_i^{(l+1)} = \mbox{UPDATE}\left(\sum_{\node_j \in \neighbor_i} \alpha_{i, j}^{(l+1)} \mes_{i,j}^{(l+1)}\right),
\end{equation}
where $\mes_{i,j}$ is the message that node $\node_j$ sends to $\node_i$, $\alpha_{i,j}$ denotes the attention weight, and $\mbox{UPDATE}\left(\cdot\right)$ indicates a function that updates the node embedding by considering messages from neighborhoods.

Based on the learned node features, a representation for a graph $\graph$ can be obtained with a set function $f_{\text{gnn}}$~(\ie, attentive pooling) as follows:
\begin{equation}
\label{eq:gnn-pool}
    \mbox{GNN}\left(\graph\right) = f_{\text{gnn}}\left( \left\{\kgout^\prime_1, \kgout^\prime_2, \dots, \kgout^\prime_n \right\} \right),
\end{equation}
where $\kgout_i^\prime$ is the embedding of node $v_i$ from the final layer. 
Like Eq. \eqref{eq:gnn-pool}, GNN-based models with message-passing perform reasoning at the node level; so, they are not readily available for path-based encoding and interpretation.
On the other hand, QAT performs reasoning at the (meta-)path level, considering a graph as a set of relations.
By tokenizing the meta-path relations, its reasoning ability is enhanced.
We verify this effect by comparing the method by which the KG is encoded, in Section~\ref{sec:analysis}.

\paragraph{RN-based methods.}
The Relation Network (RN)~\cite{santoro2017simple} is a neural network architecture that represents an object set $\Oc$ based on the relations between the objects within the set.
A general form of RN is
\begin{equation}
    \mbox{RN}\left(\Oc \right) = f\left(\left\{g\left(o_i, o_j \right)| o_i,  o_j \in \Oc \right\}\right),
\end{equation}
where function $g$ encodes the relation of the pair of objects in $\Oc$, and $f$ learns to represent the set of encoded relations. 
Considering that knowledge-based QA problems also entail representing the set of relations between concept entities, the QA problem can naturally be regarded an extension of RN.
That is, given a pair of question and answer choice, a QA problem solver encodes the set of entity relations as written in~\cite{feng2020scalable} as
\begin{equation}
\label{eq:rn_qa}
\small
    \mbox{RN}\left(\Gc \right) = f\left(\left\{g\left(\node_i, \rel, \node_j \right)| \node_i \in \nodeset_\question, \node_j \in \nodeset_\answer, \left(\node_i, \rel  ,\node_j\right) \in \edgeset \right\}\right),
\end{equation}
where function $g$~(\eg, MLP) outputs the relations between question and answer entities, and function $f$~(\eg, pooling function) aggregates the relations.
$\nodeset_\question$ and $\nodeset_\answer$ refer to the entities from the question and answer choice, respectively.

Our QAT is a more flexible and powerful architecture that generalizes the relation network.
Specifically, QAT can be formulated as follows:
\begin{equation}
\small
\begin{split}
    &\mbox{QAT}\left(\graph, \lmoutset^\text{LM} \right) \\
    &= f(\{g\left(\node_i, \rel_1, \dots, \rel_k, \node_j  \right)| \left(\node_i, \rel_1, \dots, \rel_k, \node_j \right) \in \pathset_k, 
    \\ &\quad k=1,\dots,K \} \cup \left\{\lmout_{cls}, \lmout_{1}, \dots, \lmout_{N} \right\}),
\end{split}
\end{equation}
where $\pathset_k$ indicates a set of meta-paths with length $k$.
In contrast to RN (Eq.~\eqref{eq:rn_qa}), our QAT captures the multi-hop relations (\ie, meta-paths) beyond mere one-hop relations (\ie, edges), via Meta-Path token embeddings. 
Moreover, QAT extends RN to integrate both the LMs and KGs with our relation-aware self-attention.  

%% file: 4_Experiments/4_0_experiments.tex
\input{4_Experiments/datasets}
\input{4_Experiments/implementation_details}
\input{4_Experiments/results}

%% file: 4_Experiments/datasets.tex
\subsection{Datasets}
We evaluate our method on three question-answering datasets: CommonsenseQA~\cite{talmor2019commonsenseqa}, OpenBookQA~\cite{mihaylov2018can}, and MedQA-USMLE~\cite{jin2021disease}.
Dataset statistics are in the supplement.

\paragraph{CommonsenseQA.}
This dataset contains questions that require commonsense reasoning. 
Since the official test set labels are not publicly available,
we mainly report performance on the in-house development (IHdev) and test (IHtest) sets following~\cite{lin2019kagnet}.
For CommonsenseQA, we adopt \textit{ConceptNet}~\cite{speer2017conceptnet} as the structured knowledge source.

\paragraph{OpenBookQA.}
This dataset requires reasoning with elementary science knowledge.
Here, \textit{ConceptNet} is again adopted as the knowledge graph, and we experiment on the official data split from \cite{mihaylov2018knowledgeable}.

\paragraph{MedQA-USMLE.}
This dataset originates from the United States Medical License Exam (USMLE) practice sets, requiring biomedical and clinical knowledge.
Thus, we utilize a knowledge graph provided by~\cite{jin2021disease}.
The graph is constructed via integrating the Disease Database portion of the Unified Medical Language System~\cite{bodenreider2004unified} and the DrugBank~\cite{wishart2018drugbank} knowledge graph.
We use the same data split as \cite{jin2021disease}.

%% file: 4_Experiments/implementation_details.tex
\subsection{Baselines and Experimental Setup}

We conduct experiments to compare our QAT with GNN-based methods  and RN-based methods. 
For a fair comparison, all the baselines and our model use the same LMs in all experiments.
The details on the backbone LMs and technical details are in the supplement. 


%% file: 4_Experiments/results.tex
\subsection{Experimental Results}
\noindent\textbf{CommonsenseQA.}
Here, we report QAT and its baseline performances evaluated on CommonsenseQA in Table~\ref{tab:csqa_inhouse}.
Our QAT outperforms all the methods and achieved a new SOTA in-house test accuracy (IHtest-Acc.) of 75.4\%.
The proposed model provides a 6.7\% performance boost compared to the vanilla RoBERTa language model~(RoBERTa-L (w/o KG)) that does not utilize the KG. 
Specifically, the increase over GreaseLM and JointLK suggests that our QAT's Relation-Aware Self-Attention improves joint reasoning capabilities over the LM and KG.
Also, the performance gaps compared to GSC, QA-GNN \textit{etc.} indicates the superiority of our relation-centric Meta-Path tokens.

\input{Tables/csqa_inhouse}
\input{Tables/obqa_inhouse}
\paragraph{OpenBookQA.}
In Table~\ref{tab:obqa_inhouse}, we report QAT performance on OpenBookQA.
QAT with the RoBERTa-Large renders 71.2\% mean test accuracy, which is a significant improvement over previous state-of-the-art by 0.9\% and a boost of 6.4\% compared to the vanilla RoBERTa-large LM.
Following prior works~\cite{yasunaga2021qa}, we further experiment using AristoRoBERTa as the pre-trained LM, utilizing additional text data.
By adopting AristoRoBERTa, we again achieved a state-of-the-art performance of 86.9\% mean accuracy.
This indicates that QAT successfully reasons over two different modalities and is adaptable to diverse LMs. 

\paragraph{MedQA-USMLE.}
We additionally conduct an experiment on MedQA-USMLE~\cite{jin2021disease} to evaluate the domain generality of our QAT in Table~\ref{tab:medqa}.
The table shows that QAT achieves the state-of-the-art performance of 39.3\% and improves performance over various medical LMs.
This result indicates that our QAT is effective in reasoning in various domains beyond commonsense reasoning.

%% file: Tables/csqa_inhouse.tex
\begin{table}[t!]
    \centering
    \setlength{\tabcolsep}{3.5pt}
    \begin{tabular}{l c c}
        \toprule
        \textbf{Methods} & \textbf{IHdev-Acc.}(\%) & \textbf{IHtest-Acc.}(\%) \\
        \midrule
        \midrule
        RoBERTa-L (w/o KG) & 73.1 ($\pm$0.5) & 68.7 ($\pm$0.6) \\
        \midrule
        RGCN & 72.7 ($\pm$0.2) & 68.4 ($\pm$0.7) \\
        GconAttn & 72.6 ($\pm$0.4) & 68.6 ($\pm$1.0) \\
        KagNet & 73.5 ($\pm$0.2) & 69.0 ($\pm$0.8) \\
        RN & 74.6 ($\pm$0.9) & 69.1 ($\pm$0.2) \\
        MHGRN & 74.5 ($\pm$0.1) & 71.1 ($\pm$0.8) \\
        QA-GNN & 76.5 ($\pm$0.2) & 73.4 ($\pm$0.9) \\
        CoSe-CO & 78.2 ($\pm$0.2) & 72.9 ($\pm$0.3)\\
        GreaseLM & 78.5 ($\pm$0.5) & 74.2 ($\pm$0.4) \\
        JointLK & 77.9 ($\pm$0.3) & 74.4 ($\pm$0.8)  \\
        GSC & 79.1 ($\pm$0.2) & 74.5 ($\pm$0.4) \\
        \midrule
        \textbf{QAT (ours)} & \textbf{79.5 ($\pm$0.4)} & \textbf{75.4 ($\pm$0.3)} \\
        \bottomrule
    \end{tabular}
    \caption{Performance comparison on \textit{CommonsenseQA}.}
    \label{tab:csqa_inhouse}
\end{table}



%% file: Tables/obqa_inhouse.tex
    


\begin{table}[t!]
    \centering
    \setlength{\tabcolsep}{3pt}
    \begin{tabular}{l c c}
        \toprule
        \textbf{Methods} & \textbf{RoBERTa-Large} & \textbf{AristoRoBERTa} \\
        \midrule
        \midrule
        LMs (w/o KG) & 64.8 ($\pm$2.4) & 78.4 ($\pm$1.6)  \\
        \midrule
        RGCN & 62.5 ($\pm$1.6)& 74.6 ($\pm$2.5) \\
        GconAttn& 64.8 ($\pm$1.5) & 71.8 ($\pm$1.2) \\
        RN & 65.2 ($\pm$1.2)& 75.4 ($\pm$1.4) \\
        MHGRN & 66.9 ($\pm$1.2)& 80.6 ($\pm$\text{NA}) \\ 
        QA-GNN & 67.8 ($\pm$2.8) & 82.8 ($\pm$1.6) \\
        GreaseLM & -& 84.8 ($\pm$\text{NA}) \\
        JointLK & 70.3 ($\pm$0.8) & 84.9 ($\pm$1.1) \\
        GSC & 70.3 ($\pm$0.8)& 86.7 ($\pm$0.5) \\
        \midrule
        \textbf{QAT (ours)} &\textbf{71.2 ($\pm$0.8)}    & \textbf{86.9 ($\pm$0.2)}     \\
        \bottomrule
    \end{tabular}
    \caption{Test accuracy comparison on \textit{OpenBookQA}.}
    \label{tab:obqa_inhouse}
    
\end{table}

%% file: 5_Analysis/5_0_analysis.tex
In this section, we analyze our QAT to answer the following research questions:
\textbf{[Q1]} Does each component in QAT boost performance? \textbf{[Q2]} Are relation-centric Meta-Path tokens really better than node-centric embeddings? \textbf{[Q3]} How does Relation-Aware Self-Attention utilize the language-graph relations when answering questions?

\input{Tables/medqa_usmle}
\input{Tables/analysis_ablation}
\input{Tables/analysis_gnn_ours}

\input{Tables/analysis_mp_gnn}
\paragraph{Ablation Studies.}
To answer the research question \textbf{[Q1]}, we conduct a series of ablation studies to verify the effect of each component of QAT.
By omitting Drop-MP, all the Meta-Path tokens~(MP tokens) are used in the training phase.
Ablating Cross-Modal RPB is equivalent to having no relative postion bias, and by removing MP tokens, we substitute the KG tokens with node features without additional processing.
In Table~\ref{tab:ablation}, we present the ablation results where each component is removed one by one.
By removing all our components, performance decreased significantly by 1.6\%, and the biggest degradation was observed when the MP token was ablated.
This proves the effectiveness of encoding the relation-centric information with Meta-Path tokens.

\paragraph{Meta-Path tokens vs. Node tokens.}
In order to address \textbf{[Q2]}, we elucidate the efficacy of Meta-Path tokens~\textbf{(Meta-Path)} by comparing them with \textit{node-centric} tokens.
We extensively compare our \textit{relation-centric} method with raw node-level tokens~\textbf{(Node)} as well as the GNN-processed node tokens~\textbf{(Node+GNN)}, whose GNN architecture is borrowed from \cite{yasunaga2021qa}.
Their performances are compared on CommonsenseQA and OpenBookQA in Table~\ref{tab:gnn_ours}~(above).
In all cases, we observed that the Meta-Path token consistently outperforms the node-centric tokens.

To better understand how each reacts to different question traits, we further experimented with diverse question types such as questions with negation, questions with fewer entities ($\le$ 7), and questions with more entities~($>$7).
In Table~\ref{tab:gnn_ours}~(below), our Meta-Path tokens show clear dominance over the GNN-processed node tokens.
Especially, it performs well in questions with negations by a significant margin of 3.1\%, indicating a superiority in processing logically complex questions.
Furthermore, outperformance in questions with less entities (an increase of 3.7\%) implies that our method better leverages the KG to reason over external concepts selecting the correct answer.

\input{Figures/tex/attn_choice}

\input{5_Analysis/5_0_q3.tex}

%% file: Tables/medqa_usmle.tex



\begin{table}[t!]
    \centering
    \setlength{\tabcolsep}{3.5pt}
    \begin{tabular}{l c c}
        \toprule
        \textbf{Methods} & \textbf{Accuracy} \\
        \midrule
        \midrule
        Chance                & 25.0 \\
        PMI                   & 31.1 \\
        IR-ES                 & 35.5 \\
        IR-Custom             & 36.1 \\
        ClinicalBERT-Base     & 32.4 \\
        BioRoBERTa-Base       & 36.1 \\
        BioBERT-Base         & 34.1 \\
        BioBERT-Large         & 36.7 \\
        \midrule
        SapBERT-Base~(w/o KG) & 37.2 \\
        QA-GNN                & 38.0 \\
        GreaseLM              & 38.5 \\
        \midrule
        \textbf{QAT (ours)}            & \textbf{39.3} \\
        \bottomrule
    \end{tabular}
    \caption{ Test accuracy comparison on \textit{MedQA-USMLE}.}
    \label{tab:medqa}
\end{table}

%% file: Tables/analysis_ablation.tex

\begin{table}[t]
    \centering
    \begin{tabular}{c c c | c c}
        \toprule
        \textbf{MP tokens} &\textbf{RPB}  & \textbf{Drop-MP} &  \textbf{IHtest-Acc.}(\%)\\
        \midrule
        \midrule
        \checkmark & \checkmark & \checkmark & \textbf{75.4 ($\pm$0.3)} \\    
        \checkmark & \checkmark & & 75.3 ($\pm$0.7) \\
        \checkmark &  & & 75.0 ($\pm$0.5) \\          
         & & &73.8 ($\pm$0.4)   \\
        \bottomrule
    \end{tabular}
\caption{Ablation study on \textit{CommonsenseQA}.}
    \label{tab:ablation}
\end{table}

%% file: Tables/analysis_gnn_ours.tex
\begin{table}[t!]
    \centering
    \setlength{\tabcolsep}{3.0pt}
    \begin{tabular}{l c c c}
        \toprule
        \textbf{Dataset}      &  \textbf{Node}   & \textbf{Node+GNN} & \textbf{Meta-Path} \\
        \midrule
        \midrule
        CSQA        &   73.8 ($\pm$0.4)   &  73.9 ($\pm$0.2)  &  \textbf{75.4 ($\pm$0.3)} \\
        OBQA              &  69.0 ($\pm$1.6)    &   69.6 ($\pm$0.9)   & \textbf{71.2 ($\pm$0.8)} \\
        \bottomrule
    \end{tabular}
\end{table}

%% file: Tables/analysis_mp_gnn.tex
\begin{table}[t!]
    \centering
    \small
    \begin{tabular}{l c c c}
        \toprule
        \textbf{Question Types}      &  \textbf{Node}   & \textbf{Node+GNN} & \textbf{Meta-Path} \\
        \midrule
        \midrule
        Full question set  &77.5& {77.9} & \textbf{{79.8}~($\uparrow$1.9)} \\
        Question w/ negation &75.2& {75.9} & \textbf{{79.0}~($\uparrow$3.1)} \\
        Question w/ entities $\le$ 7 &76.2& {76.2} & \textbf{{79.9}~($\uparrow$3.7)} \\
        Question w/ entities $>$ 7 &79.1& {79.5} & \textbf{{79.7}~($\uparrow$0.2)} \\        
        \bottomrule
    \end{tabular}
    \caption{Meta-Path token and Node token Comparison. The test set performances with different KG tokens are compared~(above). The development set performance on different types of questions are compared~(below).}
    \label{tab:gnn_ours}
\end{table}

%% file: Figures/tex/attn_choice.tex
\begin{figure}[t!]
\begin{center}
\includegraphics[width=1\columnwidth]{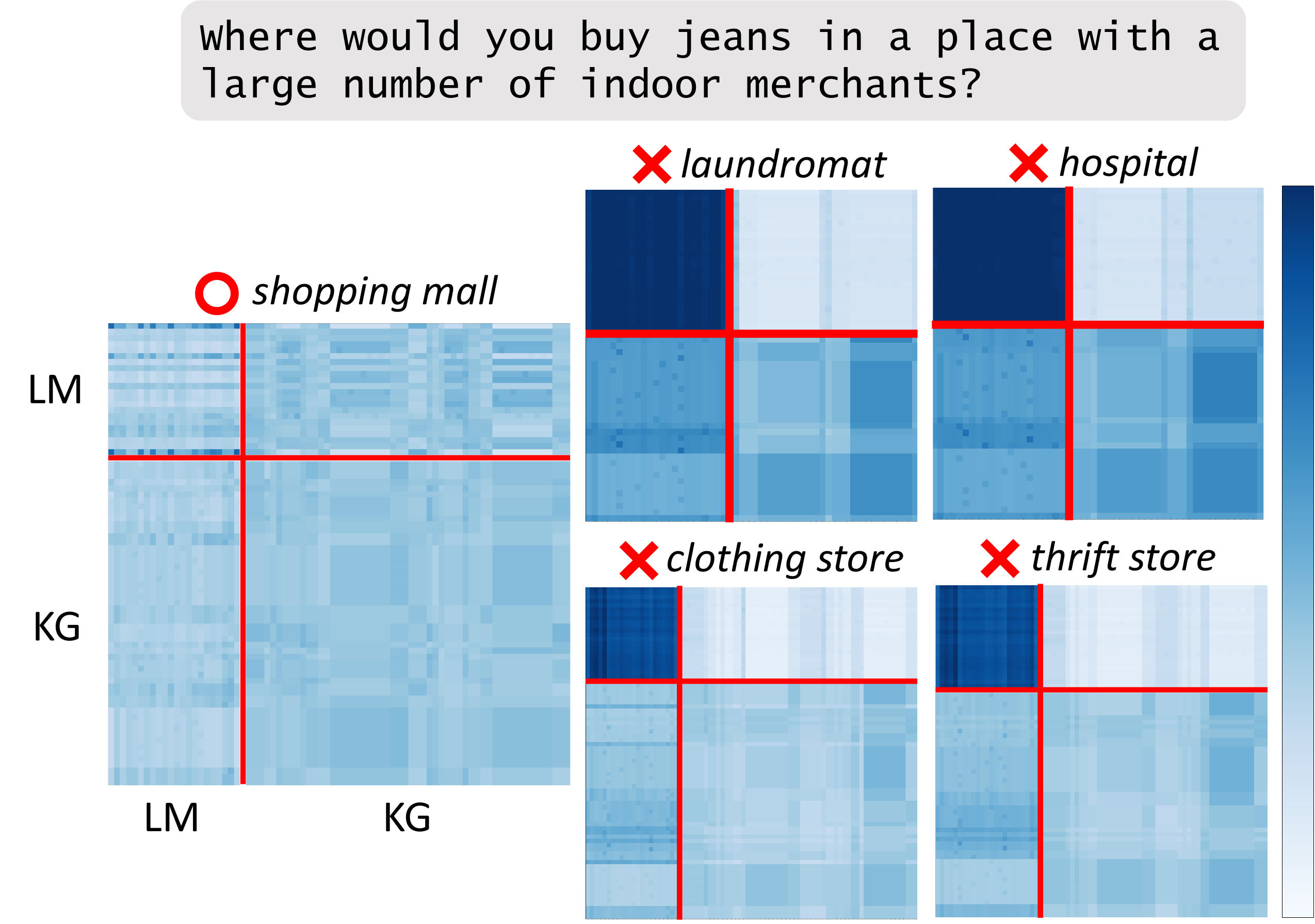}
\end{center}
\caption{ What and How QAT Selects. The attention maps for each question-answer pair is plotted. Cross-modal information is actively combined in the selected answer choice.}
\label{fig:choice}
\end{figure}

%% file: 5_Analysis/5_0_q3.tex
\paragraph{Qualitative Analyses.}
We conduct qualitative analyses to provide intuition on how QAT understands the relationship between two modalities, thereby answering \textbf{[Q3]}.
We first observe that cross-modal attention is crucial to distinguishing a correct answer from the others. 

Our Relation-Aware Self-Attention~(RASA) exhibits relatively stronger cross-modal attention, especially from KG to LM (right-upper quadrant), when a correct question-answer pair is given. 
Figure~\ref{fig:choice} shows that the attention map with the correct answer ``shopping mall'' is smooth and QAT integrates features from both modalities: LM and KG. 
On the other hand, for wrong answers (\eg, laundromat, hospital, etc.) the cross-modal attention from KG to LM (right-upper quadrant) has relatively smaller attention weights and the model mainly focuses on the LM tokens (left-upper quadrant).
These distinctive attention maps are obtained by our Meta-Path token generation. 
Note that QAT extracts a subgraph from KG for each question-answer pair and constructs a new set of Meta-Path tokens. 
Even with one-word answer choices, Meta-Path token generation results in more diversified KG token sets than node-level token generation. 
\input{Figures/tex/rpb_qual}
%

We provide additional qualitative analysis to understand how our Cross-Modal RPB further enhances cross-modal attention.
Figure~\ref{fig:quali} visualizes the attention mappings between LM and KG tokens.
To avoid clutter, we only show the strongest attention between entities, and the arrows indicate the attention direction.
For example, `left--leave' gets the most attention from `left', with RPB.
The top blue tokens, \eg, `left' and `shoes', are LM tokens, and bottom green tokens, \eg, `left--leave' and `house--entryway', are Meta-Path tokens.
Without our RPB (bottom row in Figure ~\ref{fig:quali}), we see more gray arrows indicating that cross-modal attentions between unmatched (irrelevant) tokens are prevalent.
In opposition, the RPB augments the cross-modal attention by $\omega_1$ ($LM\rightarrow KG$) and $\omega_2$ ($KG\rightarrow LM$) for matched (relevant) tokens.
Thus, matched tokens have higher attention and more red arrows are observed `with RPB' on the top row.
To be specific, (1) `house~(LM)' is selected by `house--entryway~(KG)' and `house--shoe~(KG)', (2) `house--entryway~(KG)' is selected by `entry~(LM)', `way~(LM)', and `house~(LM)' as an entity with the strongest attention.

%% file: Figures/tex/rpb_qual.tex
\begin{figure}[t!]
\begin{center}
\includegraphics[width=1\columnwidth]{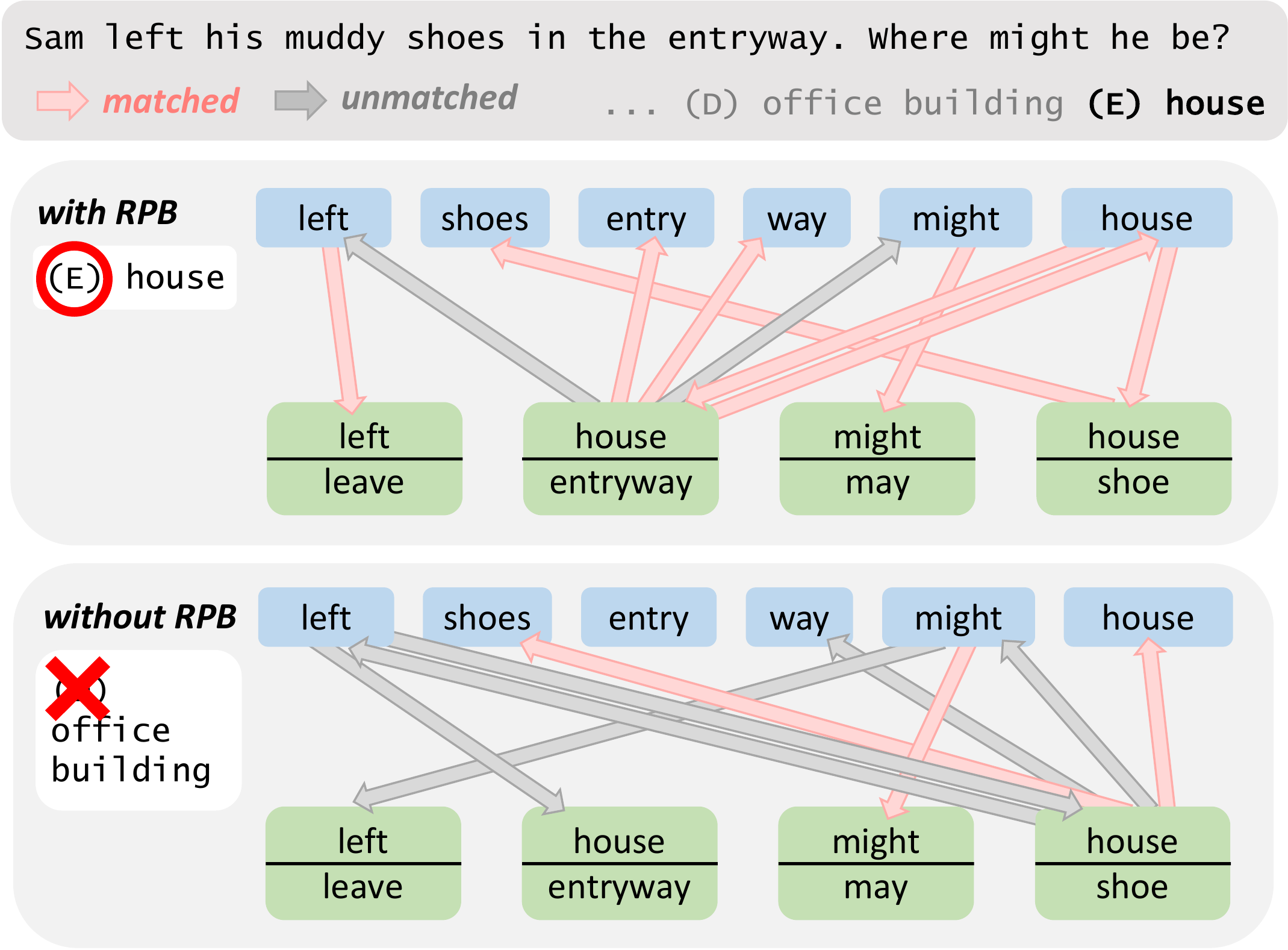}
\end{center}
\caption{The Effect of Cross-Modal Relative Position Bias. The arrows indicate the attention direction, \eg, `left--leave' gets the most attention from `left' with RPB, and the arrow colors signify whether the two tokens are matched. By removing our RPB, the attention mapping becomes un-interpretable, and leads to wrong answer selection.}
\label{fig:quali}
\end{figure}

%% file: 6_Conclusion/conclusion.tex
We proposed Question Answering Transformer~(QAT), which effectively combines LM and KG information via our Relation-Aware Self-Attention module equipped with Cross-Modal Relational Position bias.
The key to state-of-the-art performance in various QA datasets such as CommonsenseQA, OpenbookQA, and MedQA-USMLE is our Meta-Path token, which learns \textit{relation-centric} representations for the KG subgraph of a question-answer pair.

%% file: appendix.tex
\appendix

\section{Dataset Statistics}
\input{A_datasets/datasets.tex}
\section{Experimental Settings}
\input{B_experimental_settings/experimental_settings.tex}

\section{Additional Analyses}
\input{C_additional_analysis/0_analysis.tex}

%% file: A_datasets/datasets.tex
\label{sup:dataset}
\input{Tables/suppl_dataset}
We provide details on the datasets and knowledge graph we have adopted for our experiments.
In Table~\ref{tab:datasets}, we summarized the question and answer choice statistics for the three datasets we experimented on.
For CommonsenseQA~\cite{talmor2019commonsenseqa} and OpenBookQA~\cite{mihaylov2018can} datasets, we used the ConceptNet~\cite{speer2017conceptnet} as the knowledge graph.
This is a general-domain knowledge graph retaining 799,273 nodes and 2,487,810 edges in total.
Each edge is assigned to a relation type, which is merged as in Table~\ref{sup_tab:merge} following \cite{feng2020scalable,yasunaga2021qa}.
On the other hand, MedQA-USMLE~\cite{jin2021disease} requires external biomedical knowledge.
We thus used a different knowledge graph provided by~\cite{jin2021disease}, which contains 9,958 nodes and 44,561 edges.
Given each QA context, we extract a subgraph from the full knowledge graph following \cite{yasunaga2021qa}. 

%% file: Tables/suppl_dataset.tex
\begin{table}[t!]
    \centering
    \setlength{\tabcolsep}{4.0pt}
    \resizebox{1.0\columnwidth}{!}{
    \begin{tabular}{l c c}
        \toprule
        \textbf{Dataset}         & \textbf{\# Questions} & \textbf{\# Choices per Question} \\
        \midrule
        \midrule
        CommonsenseQA                      &  12,102  &  5 \\
        OpenBookQA                         &  5,957  &  4 \\
        MedQA-USMLE                        &  12,723  &  4 \\
        \bottomrule
    \end{tabular}
    }
    \caption{Dataset Statistics.}
    \label{tab:datasets}
\end{table}

%% file: B_experimental_settings/experimental_settings.tex
\label{sup:exp}
In this section, we provide details regarding our backbone language model, as well as technical details and hyperparameters we used in our experiments.

\paragraph{Language Models.}
For CommonsenseQA and OpenBookQA, we take advantage of the pretrained RoBERTa-large~\cite{liu2019roberta} model. 
In the case of OpenBookQA, we additionally apply AristoRoBERTa~\cite{clark2020f} which utilizes textual data as an external source of information.
For the MedQA-USMLE dataset, SapBERT~\cite{liu2020self}, a state-of-the-art biomedical language model, is used in place of RoBERTa models.

\paragraph{Technical Details}
\input{Tables/suppl_merge.tex}

In our experiments, the hyperparameters are tuned with respect to the development set in each dataset, and evaluated on the test set.
For training, two GeForce RTX 2080Ti or GeForce RTX TITAN GPU's were utilized in parallel.
We used RAdam as our optimizer using a linear learning rate scheduler with a warmup phase on CSQA and OBQA, and a constant learning rate on MedQA-USMLE.
Following prior works, we took the performance mean and standard deviation with three different seeds 0, 1, 2.
We kept identical hyperparameter settings across seeds and their settings vary by dataset, which are specified in Table~\ref{tab:hyperparams}.

\input{Tables/suppl_hyperparam}
\paragraph{Complex Questions.}
To validate the effectiveness of the question answering models on complex questions, we experimented with diverse question types such as questions with negation, questions with fewer entities ($\le$ 7), and questions with more entities ($>$ 7) following \cite{sun2021jointlk} in Table~5~(below) of the main paper and Table~\ref{sup_tab:complex} of the appendix.
We selected questions with negation terms by retrieving questions that contain (no, not, nothing, never, unlikely, don't, doesn't, didn't, can't, couldn't) from the CommonsenseQA IHdev set.


%% file: Tables/suppl_merge.tex
\begin{table}[t]
    \centering
    \setlength{\tabcolsep}{3.5pt}
    {
    \begin{tabular}{c c }
        \toprule
        \textbf{Relation} &\textbf{Merged Relation}\\
        \midrule
        AtLocation & \multirow{2}{*}{AtLocation} \\
        LocatedNear &                            \\
        \midrule
        Causes & \multirow{3}{*}{Causes} \\
        CausesDesire & \\
        *MotivatedByGoal & \\
        \midrule
        Antonym & \multirow{2}{*}{Antonym} \\
        DistinctFrom & \\
        \midrule
        HasSubevent & \multirow{6}{*}{HasSubevent} \\
        HasFirstSubevent & \\
        HasLastSubevent & \\
        HasPrerequisite & \\
        Entails & \\
        MannerOf & \\
        \midrule
        IsA & \multirow{3}{*}{IsA} \\
        InstanceOf & \\
        DefinedAs & \\
        \midrule
        PartOf & \multirow{2}{*}{PartOf} \\
        *HasA & \\
        \midrule
        RelatedTo & \multirow{3}{*}{RelatedTo} \\
        SimilarTo & \\
        Synonym & \\
        \bottomrule
    \end{tabular}
    }
\caption{Merged Relations. *RelationX indicates the reverse relation of RelationX.}
    \label{sup_tab:merge}
\end{table}

%% file: Tables/suppl_hyperparam.tex
\begin{table}[t!]
    \centering
    \setlength{\tabcolsep}{4.0pt}
    \small
    {
    \begin{tabular}{l| c c c}
        \toprule
        \textbf{Hyperparameter}         & \textbf{CSQA} & \textbf{OBQA} & \textbf{MedQA-USMLE} \\
        \midrule
        \midrule
        epochs                   & 30  & 75  &  50 \\
        freeze LM epochs         & 4  & 4  &  0 \\
        tolerance epochs         & 10  & 30  &  10 \\
        warmup steps         & 150  & 150  &  0 \\        
        batch size              & 128 & 128 & 128 \\
        RASA layer num.                 & 2  & 2  &  4 \\
        RASA head num.                 & 16  & 16  &  4 \\
        RASA dimension                 & 1024  & 1024  & 256  \\
        FFN dimension                 & 2048  & 2048  &  512 \\
        dropout                  & 0.1  & 0.1  & 0.1   \\
        attention dropout                 & 0.1  & 0.1  & 0.1   \\
        learning rate             & 1e-4  & 1e-4  & 1e-5  \\
        LM learning rate                & 2e-5  & 3e-5  & 6e-5  \\
        Weight decay                & 1e-2  & 1e-2  & 1e-2  \\
        Gradient norm               & 1 & 1 & 1 \\
        \multirow{2}{*}{Learning rate schedule}      & warm-up  & warm-up  & \multirow{2}{*}{constant} \\ 
        &w/ linear&w/ linear& \\
        \bottomrule
    \end{tabular}
    }
    \caption{Hyperparameter Settings.}
    \label{tab:hyperparams}
\end{table}

%% file: C_additional_analysis/0_analysis.tex
\label{sup:analysis}
In this section, we provide additional analyses to show the robustness of our QAT on diverse hyperparameter settings including regularizer coefficient, $\lambda$, and a maximum length of the path, $K$, by comparing with complex questions.
We also provide additional examples of our qualitative analyses.
\input{Tables/obqa_leaderboard.tex}
\input{C_additional_analysis/hyperparameter_sensitivity.tex}

\input{C_additional_analysis/complex_questions.tex}

%% file: Tables/obqa_leaderboard.tex
\begin{table}[t!]
    \centering
    \setlength{\tabcolsep}{3.5pt}
    \begin{tabular}{l c c}
        \toprule
        \textbf{Methods} & \textbf{Acc.} \\
        \midrule
        \midrule
        ALBERT & 81.0 \\
        AristoRoBERTa & 77.8 \\
        HGN & 81.4  \\
        AMR-SG & 81.6 \\
        ALBERT + KPG & 81.8  \\
        AristoRoBERTa + QA-GNN & 82.8 \\
        T5 & 83.2  \\
        T5 + KB & 85.4 \\
        UnifiedQA & 87.2  \\
        GreaseLM & 84.8  \\
        AristoRoBERTa + JointLK & 85.6  \\
        AristoRoBERTa + GSC & 87.4  \\
        AristoRoBERTa + QAT \textbf{(ours)} &\textbf{87.6} \\
        \bottomrule
    \end{tabular}
    \caption{\textit{OpenBookQA} official learderboard. We also report the official leaderboard performance on the test dataset.}
    \label{tab:obqa_leaderboard}
\end{table}

%% file: C_additional_analysis/hyperparameter_sensitivity.tex
\input{Tables/suppl_lambda}
\paragraph{Hyperparameter sensitivity analysis on $\lambda$.}
To evaluate our model's robustness in various hyperparameter settings, we conducted  sensitivity analyses on key hyperparameters.
Here, the magnitude of $\lambda$ is altered to see its effect on CommonsenseQA performance.
To reiterate, $\lambda$ is the constant multiplied by the Cross-Modal Relative Position Bias~(RPB) regularizer term.
As the value of $\lambda$ gets higher, the RPB scale will become larger, emphasizing the cross-modal attention between relevant tokens.
In Table~\ref{tab:lambda}, we can observe that the best test accuracy is achieved when $\lambda=10$. 
Also, we may observe that there is no dramatic performance decay even in extreme hyperparameter settings.

\input{Figures/tex/suppl_numhops.tex}
\paragraph{The effect of maximum meta-path lengths $K$.}
Here, we experimented with different maximum lengths of meta-paths, $K$, used for generating the Meta-Path tokens.
If $K=1$, only a single-hop path is considered, which is identical to using only the edges.
In Figure~\ref{sup_fig:maxlength}, we see that setting $K=2$ is optimal.
Judging from the large improvement of 0.7\% compared to $K=1$, considering multi-hop meta-path relations is crucial.
On the other hand, if the path gets too long (\ie, over $K=3$), the relational information of a meta-path gets too complex, leading to a performance decay of 0.3\%.

%% file: Tables/suppl_lambda.tex
\begin{table}[t!]
    \centering
    \setlength{\tabcolsep}{3.5pt}
    {
    \begin{tabular}{c c c}
        \toprule
        \textbf{$\lambda$}  & \textbf{IHdev-Acc}(\%) & \textbf{IHtest-Acc}(\%) \\
        \midrule
        \midrule
          0.1   & 79.3 ($\pm$0.2)     &  74.3 ($\pm$1.1)     \\
          1   &  79.0 ($\pm$0.3)  & 74.8 ($\pm$0.5)     \\
          10     & 79.5 ($\pm$0.4)&  \textbf{75.4 ($\pm$0.3)}     \\
          100   &  79.0 ($\pm$0.3)    & 75.2 ($\pm$0.6)      \\
        \bottomrule
    \end{tabular}
    }
    \caption{Sensitivity analysis on $\lambda$.}
    \label{tab:lambda}
\end{table}

%% file: Figures/tex/suppl_numhops.tex
\begin{figure}[t!]
\begin{center}
\includegraphics[width=1\columnwidth]{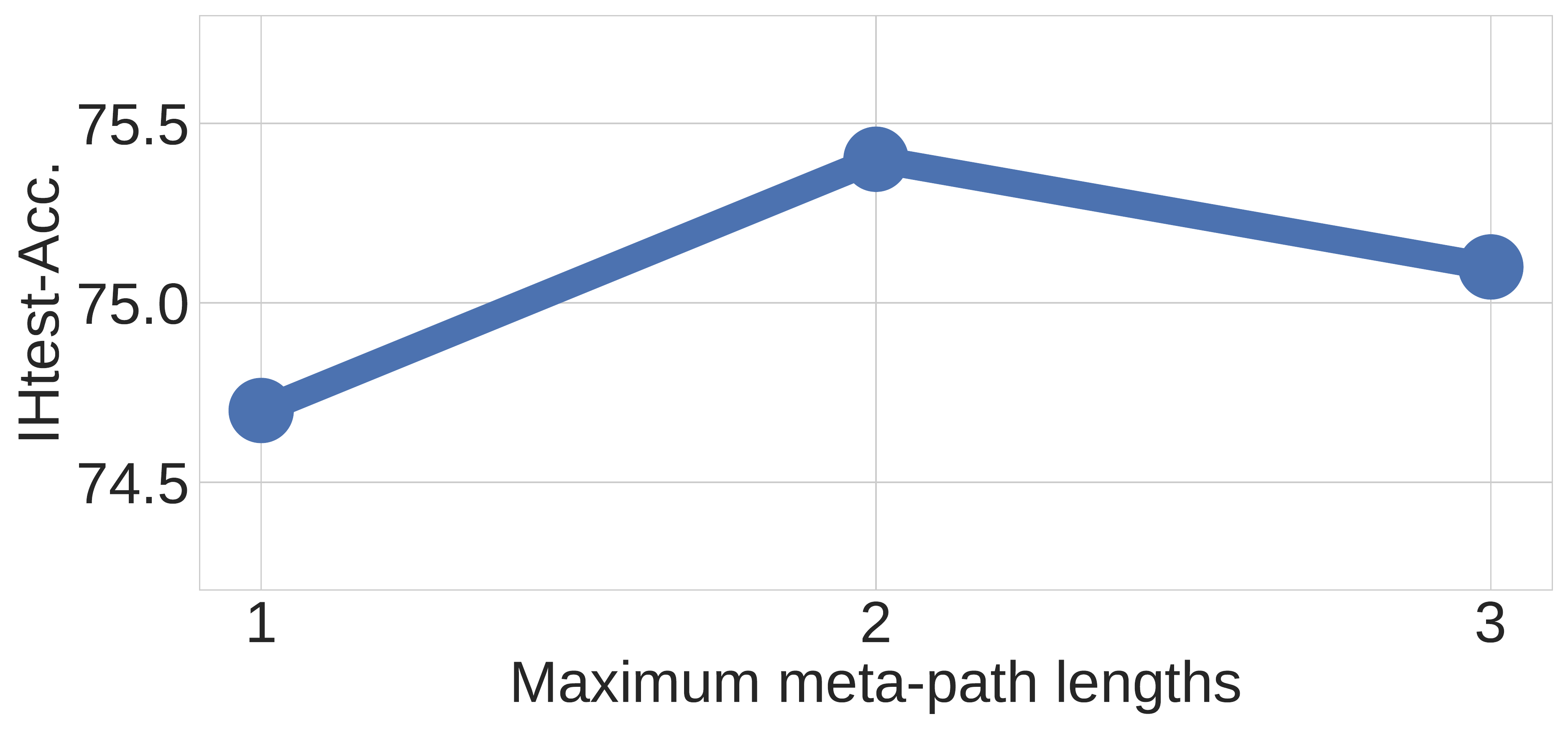}
\end{center}
\caption{The effect of maximum meta-path lengths $K$.}
\label{sup_fig:maxlength}
\end{figure}

%% file: C_additional_analysis/complex_questions.tex
\paragraph{Experimental results on complex questions.}
\input{Tables/analysis_complex}
In Table 5 of the main paper, we analyzed our method by comparing with node-centric KG encoders.
We also compare with previous works in the same settings to further validate our method's efficacy.
QAGNN~\cite{yasunaga2021qa} and JointLK~\cite{sun2021jointlk} are representative works that encoded the KG with a GNN and fused it with the LM in their unique ways.
In Table~\ref{sup_tab:complex}, our QAT tops both baseline models in all question type settings. 
We here see that our method performs especially well in questions with negation or with fewer entities, similar to our observation in the main paper.
Therefore, we can conclude that QAT is robust to logically complex questions, and makes active use of the KG for reasoning.

%% file: Tables/analysis_complex.tex
\begin{table}[t!]
    \centering
    \resizebox{1.0\columnwidth}{!}
    {
    \begin{tabular}{l c c c}
        \toprule
        \textbf{Question Types}      &  \textbf{QAGNN}   & \textbf{JointLK} &\textbf{QAT} \\
        \midrule
        \midrule
        Full question set  & {77.1} & 78.4 &\textbf{{79.8}~($\uparrow$1.4)} \\
        Question w/ negation & {72.9} & 75.2 &\textbf{{79.0}~($\uparrow$3.8)} \\
        Question w/ entities $\le$ 7 & {77.4} & 77.6 & \textbf{{79.9}~($\uparrow$2.3)} \\
        Question w/ entities $>$ 7 & {76.4} & 79.5&\textbf{{79.7}~($\uparrow$0.2)} \\        
        \bottomrule
    \end{tabular}
    }
    \caption{ \textbf{Comparison on complex question types.} }
    \label{sup_tab:complex}
\end{table}